\documentclass[letterpaper]{article} % DO NOT CHANGE THIS
\usepackage{aaai2026}  % DO NOT CHANGE THIS
\usepackage{times}  % DO NOT CHANGE THIS
\usepackage{helvet}  % DO NOT CHANGE THIS
\usepackage{courier}  % DO NOT CHANGE THIS
\usepackage[hyphens]{url}  % DO NOT CHANGE THIS
\usepackage{graphicx} % DO NOT CHANGE THIS
\usepackage{natbib}  % DO NOT CHANGE THIS AND DO NOT ADD ANY OPTIONS TO IT
\usepackage{caption} % DO NOT CHANGE THIS AND DO NOT ADD ANY OPTIONS TO IT
\usepackage{algorithm}
\usepackage{algorithmic}
\usepackage{subcaption}

\usepackage{newfloat}
\usepackage{listings}
\DeclareCaptionStyle{ruled}{labelfont=normalfont,labelsep=colon,strut=off} % DO NOT CHANGE THIS
\floatstyle{ruled}
\newfloat{listing}{tb}{lst}{}
\floatname{listing}{Listing}
\title{The Argument is the Explanation: Structured Argumentation for Trust in Agents}
\author{
    Ege Cakar\textsuperscript{\rm 1, 2},
    Per Ola Kristensson\textsuperscript{\rm 1}
}
\affiliations {
    \textsuperscript{\rm 1}Department of Engineering, University of Cambridge, Cambridge, United Kingdom\\
    \textsuperscript{\rm 2}Department of Physics, Harvard University, Cambridge, Massachusetts, USA\\
    ecakar@college.harvard.edu, pok21@cam.ac.uk
}

\begin{document}

\maketitle

\begin{center}
\textit{Preprint submitted to IAAI-26}
\end{center}
\vspace{0.5em}

\begin{abstract}
Humans are black boxes---we cannot observe their neural processes, yet society functions by evaluating verifiable arguments. AI explainability should follow this principle: stakeholders need verifiable reasoning chains, not mechanistic transparency. We propose using structured argumentation to provide a level of explanation and verification neither interpretability nor LLM-generated explanation is able to offer. Our pipeline achieves state-of-the-art 94.44 macro F1 on the AAEC published train/test split (5.7 points above prior work) and $0.81$ macro F1, $\sim$0.07 above previous published results with comparable data setups, for Argumentative MicroTexts relation classification, converting LLM text into argument graphs and enabling verification at each inferential step. We demonstrate this idea on multi-agent risk assessment using the Structured What-If Technique, where specialized agents collaborate transparently to carry out risk assessment otherwise achieved by humans alone. Using Bipolar Assumption-Based Argumentation, we capture support/attack relationships, thereby enabling automatic hallucination detection via fact nodes attacking arguments. We also provide a verification mechanism that enables iterative refinement through test-time feedback without retraining. For easy deployment, we provide a Docker container for the fine-tuned AMT model, and the rest of the code with the Bipolar ABA Python package on GitHub.
\end{abstract}

% Uncomment the following to link to your code, datasets, an extended version or similar.
% You must keep this block between (not within) the abstract and the main body of the paper.
\begin{links}
    \link{Code}{https://github.com/Ege-Cakar/Structured-Argumentation-For-Trust}
    \link{Docker Container}{egecakar/edu-classifier-serverless}
\end{links}

\section{Introduction}

Human cognitive processes remain fundamentally opaque---we cannot observe the neural mechanisms underlying our reasoning. Nevertheless, society functions by accepting this inscrutability and instead evaluating individuals based on their capacity to articulate verifiable arguments and evidence, and to guide the decision makers through their reasoning. We used to be able to more accurately pin down the internal mechanisms of AI models, yet with the advent of deep learning we have lost this insight into their internal mechanisms. As such AI models become more and more complex, and our architectures increasingly exemplify the Bitter Lesson~\cite{BitterLesson}, the goalpost for mechanistic interpretability keeps moving forward. 

We argue that in light of all this, we should treat models the way we treat humans, and require verifiable reasoning. However, with the throughput of machine learning models, having a person verify their output creates a bottleneck and raises the question of whether these models should be implemented in the first place.

As such, we require a computational approach of verifying these arguments, which we cannot find in systems that rely on similar architectures, due to stochasticity. This naturally leads us to post-hoc explainability~\cite{slack2021reliableposthocexplanations}, which analyzes decision making after the fact. However, many post-hoc explainability methods like LIME~\cite{ribeiro2016} and SHAP~\cite{shap} operate on a local scale and have explanatory power over singular outputs only. Most importantly, they don’t actually imply correctness. Thus, in light of this approach to treating machine output as we do humans, what we need is a form of source-agnostic explanatory verification, a framework that not only does not distinguish between human and machine output, but also treats the explanation \textbf{as} the verification and is only concerned with the output.

Structured Argumentation Systems (SAS's) are computational formalizations of human argumentative reasoning that have evolved over decades in traditional AI research and they provide precisely such machinery. Unlike existing Verification and Validation methods that verify against predetermined specifications, or XAI methods that solely illuminate decision processes, SAS's generate verifiable reasoning chains whose structural properties determine acceptability. Specifically, we employ Bipolar Assumption Based Argumentation (B-ABA)~\cite{A_Dickie_2025}, which extends Assumption Based Argumentation with both attack and support relations, bringing it closer to how we actually construct arguments while still providing the semantic machinery to determine acceptability and remaining computationally tractable. 

Moreover, recent advances in natural language processing that have necessitated this use of SAS's have also dissolved the historical barrier to deploying argumentation systems at scale: where previous instantiations required laborious translation of arguments into formal logical representations, contemporary NLP models, achieving near-human performance on relation classification tasks, enable natural language to serve as its own 'modal logic.' Sentences become literals, learned embeddings capture semantic relationships, and transformer-based classifiers identify argumentative relations, breathing new life into theoretical frameworks that had remained academically elegant but practically underutilized.

We exemplify the benefits of structured argumentation by tackling the problem of multi-agent AI systems for risk assessment. Risk assessment is highly labor intensive and error prone and multi-agent AI system may automate or semi-automate such risk assessment. However, a key challenge for deploying such systems is  not due to technical limitations, but because of trust issues---no reliable verification method of risk assessment outputs is currently viable.

We here present the first structured argumentation system that makes multi-agent risk assessment deployable by providing verifiable reasoning chains. Our open-source implementation runs on commodity or easily accessible hardware, democratizing access to sophisticated AI risk assessment. We make four key contributions: (1) we achieve state-of-the-art performance on AAEC argument extraction (94.44 F1) and establish a strong baseline of 0.81 F1 for 3-class relation classification on the combined AMT corpus, demonstrating that modern NLP can effectively instantiate and act as a language for structured argumentation at scale; (2) we introduce the first deployable multi-agent risk assessment system using structured argumentation for verification, addressing the trust barrier that has prevented deployment of AI risk assessment; (3) we provide a complete open-source implementation including a Docker container for easy deployment and a new Python package for Bipolar ABA; and (4) we demonstrate automatic fact-checking and iterative refinement through test-time feedback, showing how structured argumentation enables both verification and improvement without retraining.

\section{Related Work}

\subsection{Structured Argumentation}
Abstract Argumentation~\cite{O_Minh_1995}, which was introduced in 1995 to formalize and collect the research on natural language argumentation, underpins formalisms such as Assumption-Based Argumentation (ABA)~\cite{Rahwan2009} that we are utilizing a derivative of here. These systems are meant to calculate the internal consistency of arguments, and create chains of thought. In Abstract Argumentation, Bipolar extensions introduce support alongside attack ~\cite{A_Dickie_2025}, and have been captured within ABA. 
Our approach instantiates Bipolar ABA with fact nodes and SAT-backed search targeted to verification.

\subsection{Argument Mining}
Argument mining involves extracting argumentative discourse units (ADUs) and identifying their relationships (support/attack). The AAEC corpus~\cite{AnnotatedEssays} provides paragraph-level argument annotations in student essays, while the AMT corpora ~\cite{AMT1}~\cite{AMT2} offer complete within-document relations essential for document-level argumentation. Prior approaches have treated these as sequence labeling and link prediction tasks respectively~\cite{H_Potash_2017}, with recent work exploring multi-task learning~\cite{mensonides} and large language models~\cite{abkenar2024assessingopensourcelargelanguage}. However, these methods sometimes require extensive annotation, lack the computational efficiency needed for real-time deployment, or simply do not perform well.

\subsection{Trust and Verification in Multi-Agent Systems}
Recent work on trust in multi-agent AI has focused on cryptographic identity verification and runtime systems. 

The emergence of autonomous AI agents has prompted development of Zero-Trust architectures~\cite{ZeroTrustMS2025}, where agents must authenticate via OAuth2 before accessing critical functions, bringing continuous verification and least privilege principles into agentic AI. Microsoft and WSO2's collaborative work demonstrates agents carrying digital ID tokens for end-to-end auditability, but this attacks the problem of trust from a completely different angle. 

Also, recent research has shown that multi-agent LLM systems using structured communication, role-play and multiple rounds of debate can generate more trustworthy outputs~\cite{aiagents}.

However, these approaches face limitations. Cryptographic methods, most importantly, do not verify reasoning correctness, and only address trust in delegating control to LLMs, not reasoning correctness—fundamentally different from our approach. Role-play with multiple rounds of debate is helpful, which is why it is implemented here too, but still carries the same problems as trusting the output of an LLM directly. Notably absent from existing work is a framework that provides both explainability and formal verification of agent reasoning without requiring predetermined specifications—the gap our structured argumentation approach addresses.

\section{Implementation}

\begin{figure*}[h!]
    \centering
    \includegraphics[width=\textwidth]{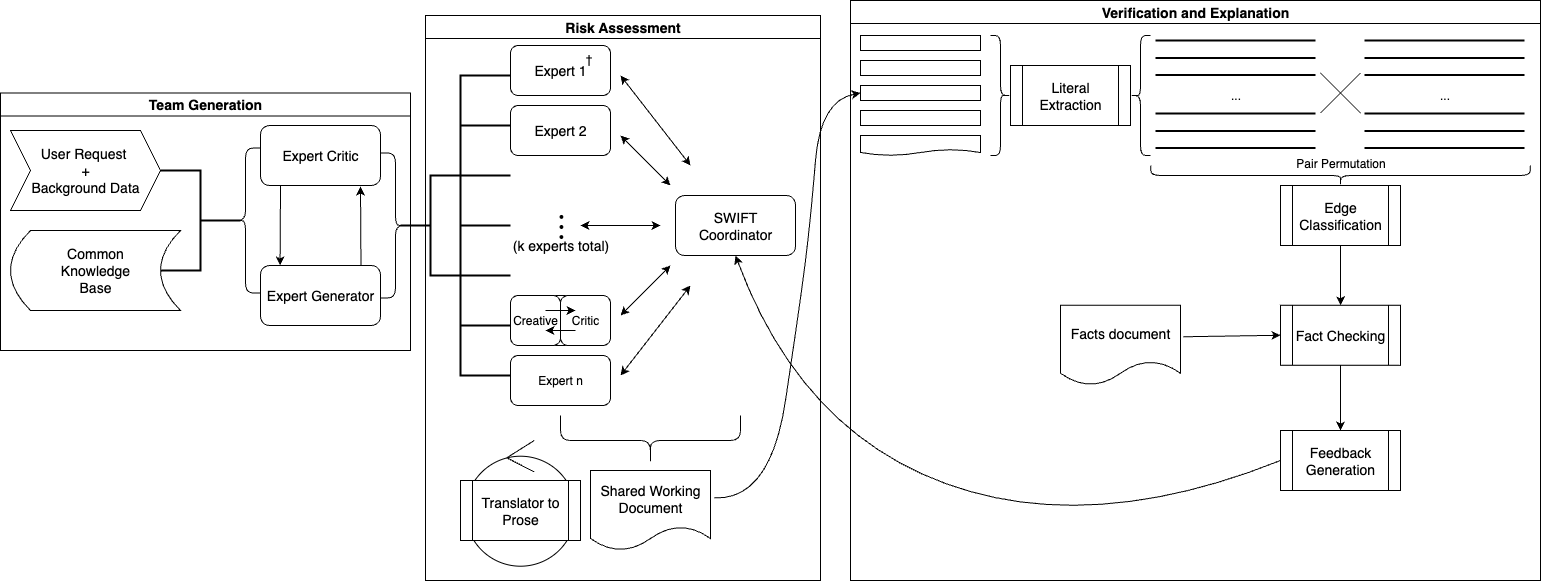}
    \caption{The complete pipeline from team generation to graph construction and verification. The SWIFT coordinator grants and receives control from experts who write to a shared working document; mining converts prose to literals, classifies relations, builds a B-ABA graph with fact nodes, and returns feedback. † All experts use a dual-agent (creative+critic) design.}
    \label{fig:framework_full}
\end{figure*}

\subsection{The Argument Mining Pipeline}

Our argument mining pipeline processes documents section by section to extract argumentative components. We support markdown and text files, where sections are determined by paragraph breaks or character limits For our literal extraction, we utilize a fine-tuned version of GPT-4.1-mini, as the costs grow linearly with input size, and GPT-4.1-mini is relatively cheap to run for great performance. The formatted dataset used to fine-tune the model is made available, so fine-tuning one for yourself is plug and play, and the settings utilized were 3 epochs, batch size = 1 and learning rate multiplier = 2. 

After all the literals in the document are extracted, we process each pair of literals with our classifier -- to make the task simpler for the classifier, we process $a,b$ and $b,a$ separately, which reduces the classification task into 3 class classification rather than 5 class classification. This means that the number of pairs to process grows with $O(n^2)$ and grows prohibitively. Thus, even though we have fine-tuned GPT 4.1 for SOTA performance on this task, we instead choose to utilize open source encoder-based models, and while we do not have official numbers for the parameter count of GPT 4.1, it is highly unlikely that it is under one trillion parameters. A ModernBERT based classifier that we fine-tuned, $\sim$ 500M parameters, can nearly match the performance of GPT 4.1, which is at least 2000$\times$ its size, and realistically even higher. 

We also provide three different extraction modes for edge classifier, within-section, window mode, and all sections. Within-section and all are just the extremes of window mode, where if we are at section $k$, and we have our window $n$, we compute all permutations of the literals within $k \pm n$, and use a set to eliminate duplicates. This allows sections to interact with each other but stops the edges from exploding too much, as it is highly unlikely that the last argument is related to the first one, for example.

We then add our fact nodes to the nodes mined, which is implemented as a \texttt{facts.md} file that gets mined and added to the graph as fact nodes. Then, we can check for any literals attacked by fact nodes, or just run our extension solvers as normal---as the conflict-freeness of extensions will take those into account.

\subsection{Bipolar Assumption-Based Argumentation}

We use Bipolar Assumption-Based Argumentation because it provides us with the suite of the necessary extensions and the mathematical foundation while not distinguishing between different types of attacks, supports, rules and literals, as well as supporting the natural notion of ``support''. If we were to utilize ASPIC+~\cite{aspic} for example, we make the classification problem much harder for our model, either turning it into a six-class classification problem, or requiring more than one model and extensive preprocessing of the text. To show the feasibility and deployability of our system, we chose to go with B-ABA. Given a set of assumptions A, contrary mappings, and rules, the framework is able to compute various extensions (admissible, preferred, stable) that represent different acceptable sets of arguments~\cite{A_Dickie_2025}. All extensions start off with the admissible property: A set that defends itself against all attacks does not attack itself and counter-attacks any external attacks. A preferred extension is a maximally admissible set, while a complete extension only includes the arguments it defends. We implement all extensions described in Dickie et al. ~\cite{A_Dickie_2025}. 
Critically, for fact verification, we restrict to unidirectional edges from fact nodes to assumptions, enabling efficient $O(n)$ verification.

We implement the entire package for B-ABA from scratch and release it as an open source Python package along with the paper so everyone can use and modify on the foundation we have built. For solving for extensions, we use Glucose~\cite{glucose2018} via turning the problem into a SAT problem, to leverage existing engineering in optimization -- we search for the $k$ largest possible admissible extensions as a pragmatic choice, as in our use case that is what we usually care about. This makes the pipeline runnable locally except for the large language models. 

\subsection{Structured What-If Technique System Setup}

To test our system, we introduce a multi-agent risk assessment system that mirrors how human expert panels operate. The system begins with intelligent team assembly. Given a risk assessment request and a user-configurable knowledge base that also comes with defaults, a generator-critic pair iteratively builds a panel of domain specialists. The generator proposes experts along with keywords for targeted knowledge retrieval, while the critic ensures comprehensive coverage without redundancy. This mimics how a project manager might assemble a diverse team, balancing expertise breadth against resource constraints. The process terminates when either sufficient coverage is achieved or resource limits (typically 5-8 experts) are reached.

Once assembled, the expert panel conducts a Structured What-If Technique (SWIFT)—a systematic risk identification  method widely used in industry~\cite{swift2012}. A coordinator agent, equipped with SWIFT methodology guidelines, orchestrates the assessment much like a human facilitator would. We also use a shared document architecture: experts can see and build upon each other's contributions, creating richer analysis than isolated assessments would produce.

Under the hood, this collaborative process follows a hub-and-spoke architecture. Each expert proposes sections for the shared document, with the coordinator determining sequence based on assessment progress and ensuring SWIFT phases are properly executed. Critically, proposed sections undergo coordinator review before integration—preventing tangential discussions while preserving valuable insights. This mirrors how effective human panels operate, with a facilitator maintaining focus while encouraging diverse perspectives.

Each expert itself employs a dual-agent design inspired by human deliberative processes, akin to how we sometimes have internal debates. A creative agent proposes ideas while a critical agent filters for relevance and feasibility. A summary agent then consolidates these deliberations into coherent document sections. This architecture produces more nuanced, self-corrected assessments than single-agent approaches.

Finally, since downstream argument mining models perform optimally on natural prose rather than structured data, an asynchronous translator converts any tabular content into flowing text while preserving any content that is already in natural text. This preprocessing ensures compatibility with our mining pipeline without information loss or extensive tampering. The output is a comprehensive risk assessment document, available both as structured markdown and segmented sections for subsequent analysis. We argue that the deployment of such a system without our verification pipeline would be suboptimal due to trust issues with stakeholders. 

\section{Evaluation}

\subsection{Literal Extraction}

In our framework, literals are atomic argumentative discourse units extracted from text: essentially claims or premises that serve as nodes in the argument graph. For literal extraction, we fine-tune GPT-4.1-mini and GPT-4.1 on the AAEC corpus~\cite{gpt41}. We fine-tune for 3 epochs, with a batch size of 1 and learning rate multiplier of 2 on the published essay-level train/test split from the AAEC corpus with seed = 28 ~\cite{AnnotatedEssays}. We evaluate untyped argumentative span extraction (claims and premises merged). Under strict exact-span matching, our model achieves F1 = 86.97.

Prior AAEC work reports various token-level scores using different evaluation protocols. On the published train/test split, Mensonides et al.~\cite{mensonides} achieve 88.70 F1 using multi-task deep learning with auxiliary tasks for argument component detection. Ajjour et al.~\cite{ajjour-etal} report 88.54 macro-F1 on the Eger split (a different essay-level split). Methods using different evaluation setups that can take advantage of more data and less classes report higher scores, such as Demaria et al.~\cite{demaria} achieving 91.05 micro F1 using 5-fold cross-validation on binary in/out classification and only 87.33 micro F1 on BIO classification.

We achieve 94.44 token-F1 on the published split using BIEO tagging, exceeding the best comparable result (Mensonides et al.) by 5.7 points without requiring auxiliary tasks. Our exact-span F1 of 86.97 constitutes a new benchmark for this evaluation protocol. To our knowledge, this is the first dedicated evaluation of argument boundary detection as a four-class sequence-labeling task on the AAEC corpus, focusing solely on span identification without role classification—a foundational component for structured argumentation systems like B-ABA that do not distinguish between claims and premises. Table \ref{tab:aaec_results} is a comparison of different results.

We chose GPT-4.1 over smaller models to avoid dataset overfitting, following Feger ~\cite{L_Feger_2025}. The model's size and pre-trained language understanding make it unlikely to merely memorize dataset patterns when fine-tuned with LoRA. Given that inference cost scales linearly $O(n)$ with text length, the superior generalization justifies the higher API costs.

\begin{table}[th]
\centering
\caption{Argument boundary detection on AAEC. Our models use the published essay-level train/test split from the AAEC corpus. Our per-class F1 values are available in the appendix.}
\label{tab:aaec_results}
\resizebox{\columnwidth}{!}{%
\begin{tabular}{|l|c|c|c|c|}
\hline
\textbf{Method} & \textbf{Task} & \textbf{Token-F1} & \textbf{Exact-F1} & \textbf{IoU} \\
\hline
Ajjour et al.\ (2017)     & BIO                    & 88.54   & --     & --    \\
Mensonides et al.\ (2019)$^{*}$        & BIO              & 88.70   & --     & --    \\
Demaria et al.\ (2022)$^{\dagger}$     & Binary (A vs O)        & 91.05   & --     & --    \\
\hline
\textbf{GPT-4.1-mini (Ours)}           & BIEO                   & 94.08 & 86.21 & 94.97 \\
\textbf{GPT-4.1 (Ours)}                & BIEO                   & \textbf{94.44} & \textbf{86.97} & \textbf{94.88} \\
\hline
\end{tabular}%
}
\par\footnotesize
$^{*}$ Published train/test split, multi-task learning with POS/chunking auxiliary tasks. \\
$^{\dagger}$ 5-fold CV; binary token-level F1 (A vs O), typically yields higher scores. \\
\end{table}

The model processes text sections and outputs a JSON dictionary mapping literal IDs to extracted text spans. For evaluation against baselines, we convert these spans back to token-level labels.

\subsection{Relation Classification}

We evaluate relation classification on the Argumentative MicroTexts (AMT) corpus, chosen over AAEC as it provides all argumentative relations inside a document, essential for document-level argumentation mining, as opposed to AAEC only providing within paragraph edges. While AMT originally annotates only support and attack relations, we extend it to a 3-class task by generating neutral pairs from non-connected argument components within the same documents, creating a more realistic classification scenario with class distribution of approximately 40\% support, 40\% neutral, and 20\% attack.

We use the combined AMT corpus (Parts 1 and 2, totaling $\sim$ 290 texts) to maximize training data and real-world applicability. While Part 1 consists of 112 controlled, professionally translated texts, Part 2 adds 171 crowd-sourced texts with potentially noisier annotations. This combination presents trade-offs: increased annotation inconsistency but 2.5$\times$ more training examples, particularly beneficial for data-hungry transformer models. The neutral pairs are generated by sampling argument component pairs that have no annotated relation in the original corpus, reflecting real-world scenarios where most argument pairs lack direct relationships.

Our goal was not to optimize for benchmark comparison but to develop a robust component for our end-to-end argumentation framework. We therefore chose to use all available data to maximize real-world applicability, achieving strong performance (0.81 F1) that exceeds published baselines while maintaining efficiency through our 500M parameter ModernBERT-large model that achieves near same performance (0.79 F1) compared to GPT-4.1's 1T+ parameters.

While Potash et al. (2017) achieve 0.74 F1 on the related but distinct task of binary link prediction, we are not aware of prior work reporting comparable results on the 3-class relation classification task (support/attack/neutral) with gold ADU boundaries on AMT. Recent work by Cabessa et al. achieves 0.747 F1 using LLaMA-3-70B on AbstRCT (average of 3 test sets)~\cite{cabessa-etal-2025-argument} on the comparable ARIC task (combined task of determining whether two arguments are related, and if so, attack/support). While these are not directly comparable due to dataset differences, our approach achieves 0.79 \& 0.81 F1 on AMT, demonstrating strong performance on this challenging 3-class classification task. Table \ref{tab:amt_relation_results} is a comparison of results.

We have deployed our ModernBERT model for the pipeline due to its performance comparable to much larger models, as encoders just seem to be a better architecture for this task. 

\begin{table}[th]
\centering
\caption{Relation classification on AMT corpus. We extend the original 2-class task (support/attack) to 3-class by generating neutral pairs from non-connected ADUs. All our models use combined Parts 1+2 (~290 texts). We report macro F1 throughout for transparent evaluation on imbalanced classes. Detailed results in the appendix.}
\label{tab:amt_relation_results}
\resizebox{\columnwidth}{!}{%
\begin{tabular}{|l|c|c|c|c|}
\hline
\textbf{Method} & \textbf{Classes} & \textbf{Data} & \textbf{Macro-F1} & \textbf{Params} \\
\hline
Potash et al.\ (2017)$^{\dagger}$        & 2 & Part 1     & 74.0   & -- \\
Abkenar et al.\ (2024)$^{*}$ - Mistral   & 2 & Part 1     & 65.1$^{\mu}$  & 7B \\
Abkenar et al.\ (2024)$^{*}$ - Mixtral   & 2 & Part 2     & 73.4$^{\mu}$  & 8×7B \\
Abkenar et al.\ (2024)$^{*}$ - Llama3    & 2 & Part 2     & 71.9$^{\mu}$  & 8B \\
\hline
\multicolumn{5}{|l|}{\textit{Few-shot In-Context Learning (5-shot)}} \\
\hline
\textbf{GPT-4.1 (Ours)}                  & 3 & Parts 1+2  & 71.0   & estimated $>$1T \\
\textbf{Claude Sonnet 4 (Ours)}          & 3 & Parts 1+2  & 72.3   & -- \\
\hline
\multicolumn{5}{|l|}{\textit{Fine-tuned Models}} \\
\hline
\textbf{Gemini 2.5 Pro (Ours)}           & 3 & Parts 1+2  & 74.1   & -- \\
\textbf{ModernBERT-large (Ours)}         & 3 & Parts 1+2  & 79.0   & 0.5B \\
\textbf{GPT-4.1 (Ours)}                  & 3 & Parts 1+2  & \textbf{81.0}   & $>$1T \\
\hline
\end{tabular}%
}
\par\footnotesize
$^{\dagger}$ Pointer networks on Part 1 only (112 texts); 2-class task (support/attack). \\
$^{*}$ Reported as ``zero-shot'' but includes examples in system prompt (few-shot by standard definition). \\
$^{\mu}$ Micro F1 reported; typically higher than macro F1 on imbalanced data. \\
Note: Our 3-class task includes neutral pairs generated from non-connected ADUs within documents.
\end{table}

\subsection{System Demonstration}

\begin{figure}[t]
    \centering
    \begin{subfigure}[b]{0.48\columnwidth}
        \centering
        \includegraphics[width=\textwidth]{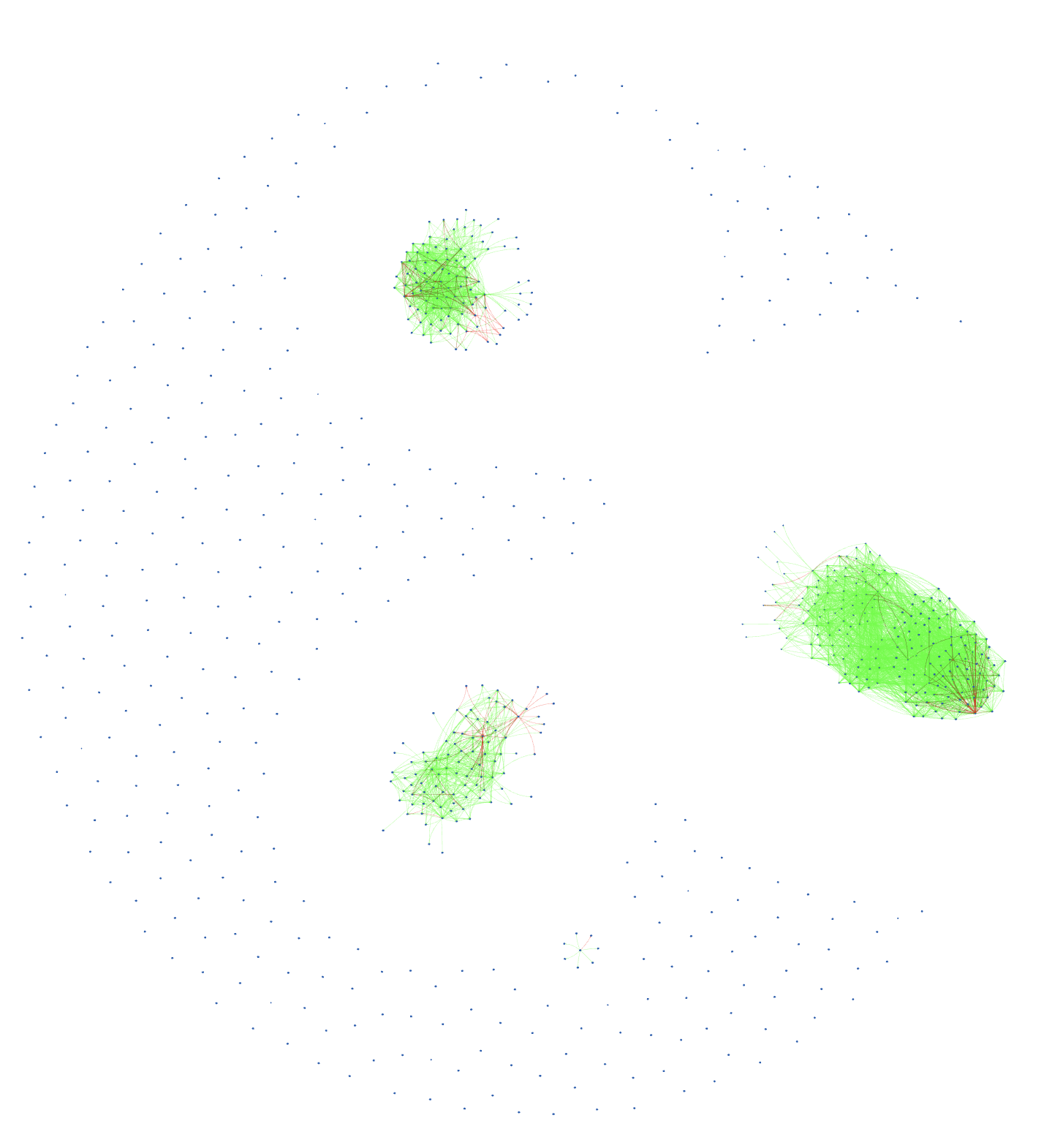}
        \caption{The sparse attack edges (1:12 ratio) demonstrate the collaborative nature of risk assessment documents where arguments build upon rather than contradict each other.}
        \label{fig:bare_graph}
    \end{subfigure}
    \hfill
    \begin{subfigure}[b]{0.48\columnwidth}
        \centering
        \includegraphics[width=\textwidth]{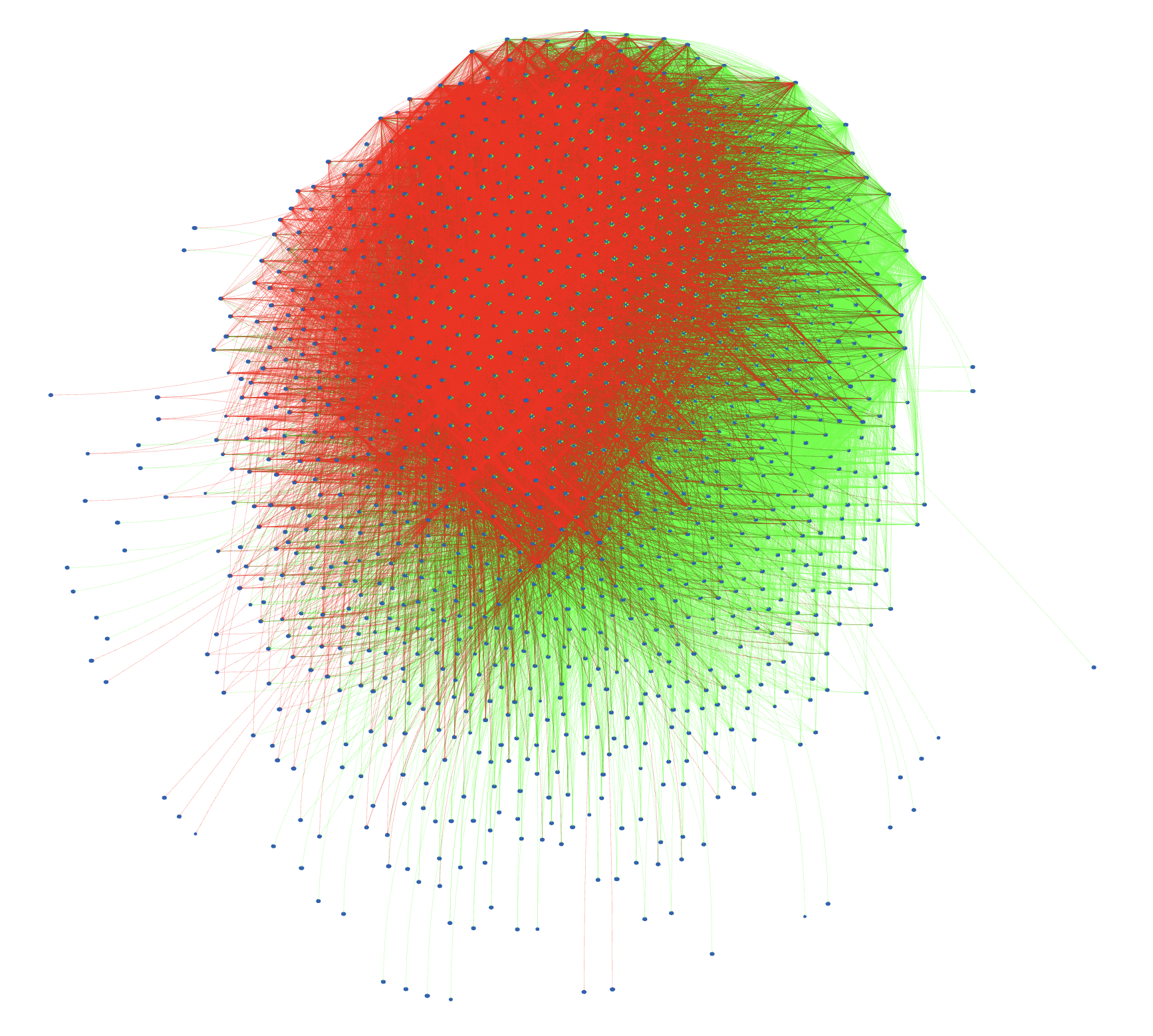}
        \caption{Obama-McCain Debate Graph. The higher proportion of red attack edges (1:4 ratio) reflects the adversarial nature of political debate, validating our classifier's ability to detect varying argumentative structures across different document types.}
        \label{fig:debate}
    \end{subfigure}
    \caption{Comparison of argumentation graphs from different document types. The edge colors (green=support, red=attack) reveal the distinct argumentative patterns: collaborative in technical documents vs. adversarial in debates.}
    \label{fig:graph_comparison}
\end{figure}

Figure \ref{fig:bare_graph} is the graph of a risk assessment generated by the multi-agent system. An immediate observation is the concentration of support edges as opposed to attack edges -- this makes sense, since risk assessment reports typically build upon prior content. Not only that, just due to the nature of LLMs, the writing that came before affects their output. In this case, the agents were explicitly instructed to dissect what came before them and were told to disagree if it did not match up with their ``expertise,'' which they got from the database with their unique keywords. 

Due to the nature of our documents then, as well as the nature of the application focus where larger consistent sets of arguments are preferable to smaller ones, our solver starts from larger sets then iteratively removes nodes. This enables us to find extensions in reasonable times fully locally. 

To test out the hypothesis that the support-heaviness was due to the document structure and not a fault of the edge classifier, we also ran the same pipeline on something we know has a lot more attack edges---the first 2008 Obama-McCain debate~\cite{obama_mccain_2008}. The results are clear in figure \ref{fig:debate}.

\subsubsection{Fact Checking} 
\begin{figure}[t]
    \centering
    \includegraphics[width=0.5\linewidth]{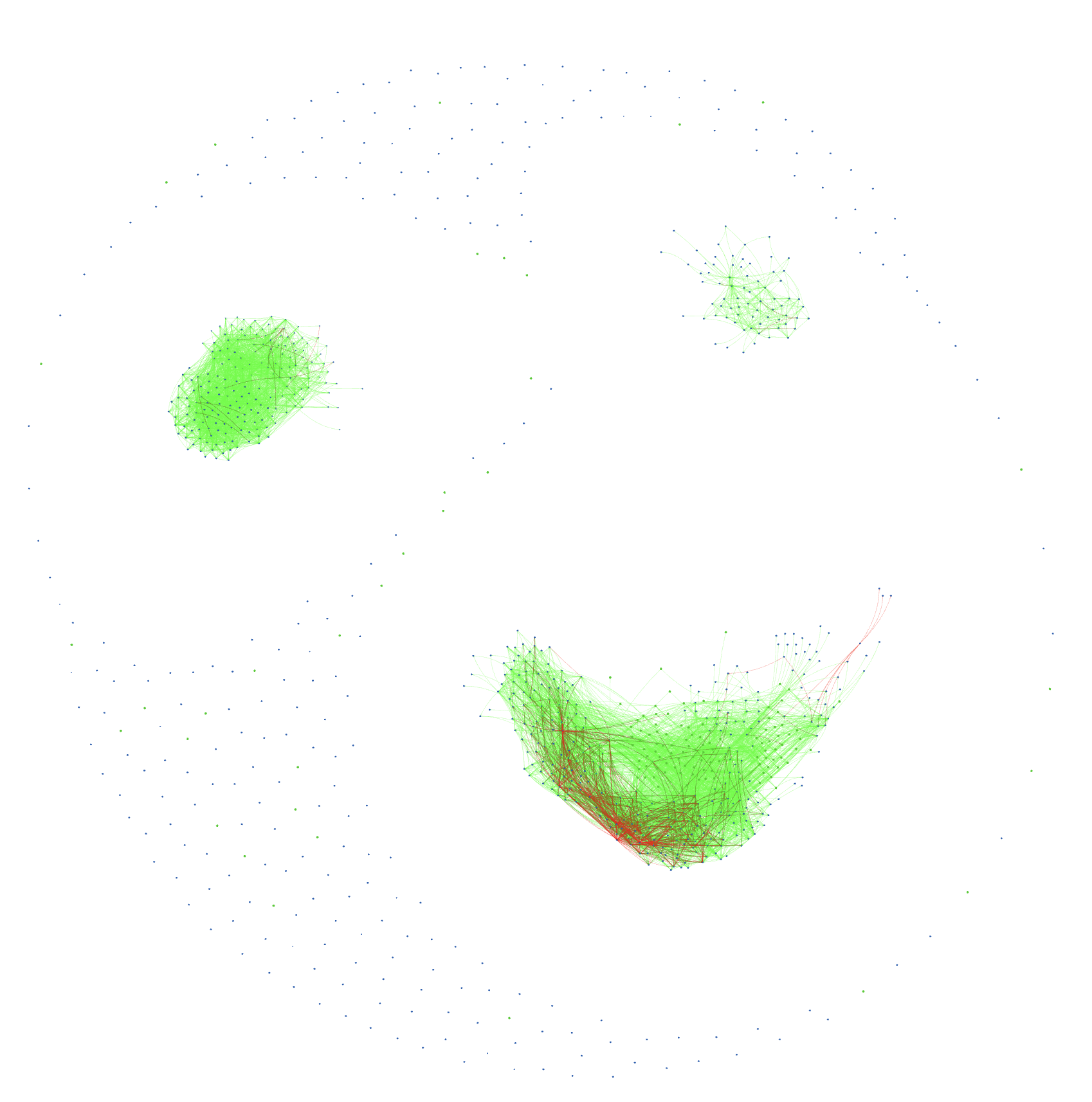}
    \caption{The risk assessment graph after fact-checking. Fact nodes inject attack edges into the structure, creating clusters where factual contradictions are detected. Approximately 3x the baseline attack rate validates the automatic verification capability.}
    \label{fig:fact_graph}
\end{figure}

The implementation takes a saved graph file as well as a facts.md, extracts the literals from facts.md, and checks edges from facts to assumptions. We are then able to call a method from the Bipolar ABA package that shows each literal attacked by a fact, as well as which fact is attacking it. This system is also able to encode for supporting and conflicting preferences as well as we only need the relation classifier to be able to detect those, which our classifier has demonstrated capability in. Our framework was able to pinpoint attack edges from the facts we have planted in the \texttt{facts.md} file that were chosen to be contradictory to statements chosen by hand from the risk assessment document. The same risk assessment document from figure \ref{fig:bare_graph} is fact-checked in figure \ref{fig:fact_graph}. The graphs can be analyzed interactively using the GitHub repository. 

For transparency: in a real world use case, these lists of facts would be provided in the initial prompt by the user. Here, we chose not to do so, to increase the number of matches and demonstrate the system. 

\subsubsection{Feedback Loop} 
For the feedback loop, we utilize our existing checkpointing system. We simply save all messages ever sent (except for internal debate messages), and when we want to provide feedback, go back to the latest checkpoint file, attach a message from the coordinator explaining the feedback it got from the user, and let them address those. 

The way we do argument checking is that we first take every argument that has been fact-checked, and provide those, including the reasoning chain that led to them (truncated, since the structure is support heavy), and the facts. 

We also identify key literals that should be well-supported and check if they face undefended attacks within depth $m$. When found, we provide feedback containing the attack chain, weak links, and attacking arguments. For example replies, see appendix, as the models are very verbose.

\section{Limitations and Future Work}

While our implementation demonstrates strong empirical performance, we identify several theoretical and practical directions for development. We modified existing Bipolar ABA theory for our use case, particularly with unidirectional fact edges and fact-checking methods. Though empirically effective, a more formally grounded framework could be developed. In support-dominant networks like risk assessments, admissibility computations become challenging as closures propagate attacks across extreme distances. Limiting depth helps feasibility but better-suited theoretical foundations could improve this.

The relation classifier remains the most important improvement avenue. While achieving 0.79-0.81 F1, a larger encoder model trained on more diverse data could overcome dataset-specific learning issues. Our ModernBERT model handles only 512 tokens despite supporting 8192, as no training pairs exceeded this length. Synthetic augmentation proved ineffective as models learned generator artifacts rather than argumentation patterns. Training on NLI datasets as auxiliary tasks or theorem-proving corpora could provide substantial improvements.

Literal extraction could benefit from longer, more self-contained argumentative units. Providing surrounding context to the classifier could improve accuracy. While we used decoder models for their availability, large encoder models with token-level classification might yield better extraction. The feedback loop, though functional, needs optimization for choosing which attacked literals to defend out of the ones getting attacked and determining optimal argument chain depths.

Our open-source implementation enables immediate deployment, with the complete pipeline running on commodity hardware. The system applies to LLM-based search engines, fact-checking, and domains where no one currently deploys multi-agent systems at scale. Future work should explore automatic fact node generation, local literal extraction models, and applications to other verification-critical domains while maintaining the computational efficiency that makes our approach practical.

\section{Conclusion}

In this paper we presented a deployable structured argumentation system that advances trust and verification in multi-agent AI. Our approach achieved new horizons on established benchmarks, 94.44 F1 on AAEC literal extraction and 0.81 F1 on AMT relation classification, while demonstrating practical verification capabilities through Bipolar ABA. We developed and deployed the first multi-agent risk assessment system using structured argumentation, addressing the trust barrier that has prevented widespread adoption of AI risk assessment. The system employed specialized agents collaborating through the Structured What-If Technique, with their outputs converted into verifiable argument graphs. Using Argumentation, we captured support and attack relationships, enabling automatic fact-checking through unidirectional edges from fact nodes. We also provided iterative refinement through test-time feedback. Our open-source implementation, Docker container for the relation classifier and complete Python package for Bipolar ABA enables immediate deployment on commodity hardware. We demonstrated that structured argumentation can provide the verification guarantees needed for deploying multi-agent AI systems in critical applications, offering a practical path forward for trustworthy AI deployment. This work was submitted to the Thirty-Eighth Annual Conference on Innovative Applications of Artificial Intelligence (IAAI-26).

\clearpage

\appendix
\section*{Appendix}

\subsection{ModernBERT Training Details}

The largest problem in training the ModernBERT model was the class imbalance. For this, we tried many different methods, in the end settling on label smoothing. Utilizing label smoothing = 0.05, with a learning rate of $10^{-5}$ and batch size of 4, we were able to reach our results of $\approx 0.8$. Architecturally, we simply utilized the ModernBERT weights as our encoder, and trained a new classification head. We did full parameter fine-tuning, meaning the encoder-weights were also updated.

\subsection{Detailed Results}

\begin{table}[h]
\centering
\caption{GPT-4.1 Performance on AMT Relation Classification}
\begin{tabular}{|l|c|c|c|c|}
\hline
\textbf{Class} & \textbf{Prec.} & \textbf{Rec.} & \textbf{F1} & \textbf{N} \\
\hline
Support & 0.817 & 0.798 & 0.807 & 168 \\
Rebuttal & 0.759 & 0.837 & 0.796 & 49 \\
None & 0.829 & 0.825 & 0.827 & 217 \\
\hline
\textit{Macro} & 0.802 & 0.820 & \textbf{0.810} & -- \\
\textit{Micro} & 0.816 & 0.816 & \textbf{0.816} & 434 \\
\hline
\end{tabular}
\end{table}

\begin{table}[h]
\centering
\caption{ModernBERT-large Performance on AMT Relation Classification (seed=42)}
\begin{tabular}{|l|c|c|c|}
\hline
\textbf{Class} & \textbf{Precision} & \textbf{Recall} & \textbf{F1 Score} \\
\hline
Support & 0.827 & 0.827 & 0.827 \\
Rebuttal & 0.889 & 0.696 & 0.780 \\
None & 0.756 & 0.813 & 0.782 \\
\hline
\textit{Macro} & 0.824 & 0.778 & \textbf{0.796} \\
\textit{Micro} & 0.802 & 0.802 & \textbf{0.802} \\
\hline
\end{tabular}
\end{table}

The model appears conservative in predicting rebuttals, but rebuttal guesses have good performance. 

\begin{table}[h]
\centering
\caption{GPT-4.1 Literal Extraction Performance on AAEC}
\begin{tabular}{|l|c|c|c|}
\hline
\textbf{Evaluation Metric} & \textbf{Precision} & \textbf{Recall} & \textbf{F1/Score} \\
\hline
Exact Span (Strict) & 0.881 & 0.861 & 0.870 \\
Word-Level (Macro) & -- & -- & 0.944 \\
\quad start\_arg & -- & -- & 0.919 \\
\quad in\_arg & -- & -- & 0.969 \\
\quad end\_arg & -- & -- & 0.946 \\
IoU (Average) & -- & -- & 0.949 \\
\hline
\end{tabular}
\end{table}

\begin{table}[h]
\centering
\caption{GPT-4.1-mini Literal Extraction Performance on AAEC}
\begin{tabular}{|l|c|c|c|}
\hline
\textbf{Evaluation Metric} & \textbf{Precision} & \textbf{Recall} & \textbf{F1/Score} \\
\hline
Exact Span (Strict) & 0.875 & 0.852 & 0.862 \\
Word-Level (Macro) & -- & -- & 0.941 \\
\quad start\_arg & -- & -- & 0.907\\
\quad in\_arg & -- & -- & 0.969 \\
\quad end\_arg & -- & -- & 0.946 \\
IoU (Average) & -- & -- & 0.950 \\
\hline
\end{tabular}
\end{table}

As it can be seen, 4.1 mini and 4.1 get essentially the same performance in literal extraction. 

\subsection{Deployment Examples}

 Figure \ref{fig:coordinator-feedback} is a sample response from the coordinator after the feedback is provided. 

\begin{figure}[h]
\centering
\fbox{\parbox{0.95\columnwidth}{
\small
\textbf{Coordinator Response to Feedback:}\\[0.5em]
\textit{``The auto-generated critique targets our weakest leverage points: timebase integrity, event ordering, latency tails, and cold-start behavior. Addressing these first will harden multiple attacked assumptions at once (e.g., a458, a494, a512), while clarifying safe-state behavior when clocks drift or compute resets occur (a441) and preventing test-mode bleed-through (a460).}\\[0.5em]
\textit{The Real-Time Edge Computing \& Performance Assurance Expert is best suited to produce a Step 8 Addendum that adds enforceable, testable controls for monotonic ordering, PTP grandmaster failover, hard maximum latency with watchdog-enforced safe-state, and JIT warmup of models/GPUs.}\\[0.5em]
\textit{This supports prior Step 8 latency and precedence requirements; however, it challenges any reliance on QoS/segmentation alone and any absence of monotonic decision gating or cold-start controls. Next turn I will QC and attempt merge.''}
}}
\caption{Sample coordinator response after receiving fact-checking feedback, demonstrating the system's ability to identify and prioritize weaknesses in the argumentation structure, as well as coordination performance..}
\label{fig:coordinator-feedback}
\end{figure}

\bibliography{aaai2026}

\begin{thebibliography}{24}
\providecommand{\natexlab}[1]{#1}

\bibitem[{Abkenar et~al.(2024)Abkenar, Wang, Graupner, and Stede}]{abkenar2024assessingopensourcelargelanguage}
Abkenar, M.~Y.; Wang, W.; Graupner, H.; and Stede, M. 2024.
\newblock Assessing Open-Source Large Language Models on Argumentation Mining Subtasks.
\newblock arXiv:2411.05639.

\bibitem[{Ajjour et~al.(2017)Ajjour, Chen, Kiesel, Wachsmuth, and Stein}]{ajjour-etal}
Ajjour, Y.; Chen, W.-F.; Kiesel, J.; Wachsmuth, H.; and Stein, B. 2017.
\newblock Unit Segmentation of Argumentative Texts.
\newblock In Habernal, I.; Gurevych, I.; Ashley, K.; Cardie, C.; Green, N.; Litman, D.; Petasis, G.; Reed, C.; Slonim, N.; and Walker, V., eds., \emph{Proceedings of the 4th Workshop on Argument Mining}, 118--128. Copenhagen, Denmark: Association for Computational Linguistics.

\bibitem[{Audemard and Simon(2018)}]{glucose2018}
Audemard, G.; and Simon, L. 2018.
\newblock On the Glucose SAT solver.
\newblock \emph{International Journal on Artificial Intelligence Tools}, 27(1): 1--25.
\newblock Hal-03299473.

\bibitem[{Cabessa, Hernault, and Mushtaq(2025)}]{cabessa-etal-2025-argument}
Cabessa, J.; Hernault, H.; and Mushtaq, U. 2025.
\newblock Argument Mining with Fine-Tuned Large Language Models.
\newblock In Rambow, O.; Wanner, L.; Apidianaki, M.; Al-Khalifa, H.; Eugenio, B.~D.; and Schockaert, S., eds., \emph{Proceedings of the 31st International Conference on Computational Linguistics}, 6624--6635. Abu Dhabi, UAE: Association for Computational Linguistics.

\bibitem[{Card, Ward, and Clarkson(2012)}]{swift2012}
Card, A.~J.; Ward, J.~R.; and Clarkson, P.~J. 2012.
\newblock Beyond FMEA: the structured what-if technique (SWIFT).
\newblock \emph{Journal of Healthcare Risk Management}, 31(4): 23--29.

\bibitem[{{Commission on Presidential Debates}(2008)}]{obama_mccain_2008}
{Commission on Presidential Debates}. 2008.
\newblock General Election Presidential Debate.
\newblock \url{https://www.debates.org}.
\newblock First presidential debate between John McCain and Barack Obama. University of Mississippi, Oxford, MS. Moderated by Jim Lehrer (PBS). Transcript available at debates.org.

\bibitem[{de~Cerqueira et~al.(2025)de~Cerqueira, Agbese, Rousi, Xi, Hamari, and Abrahamsson}]{aiagents}
de~Cerqueira, J. A.~S.; Agbese, M.; Rousi, R.; Xi, N.; Hamari, J.; and Abrahamsson, P. 2025.
\newblock Can We Trust AI Agents? A Case Study of an LLM-Based Multi-Agent System for Ethical AI.
\newblock arXiv:2411.08881.

\bibitem[{Demaria et~al.(2022)Demaria, Delsanto, Colla, Mensa, Pasini, and Radicioni}]{demaria}
Demaria, R.; Delsanto, M.; Colla, D.; Mensa, E.; Pasini, E.; and Radicioni, D.~P. 2022.
\newblock Shuffling-Based Data Augmentation for Argument Mining.
\newblock In Confalonieri, R.; and Porello, D., eds., \emph{Proceedings of the 6th {Workshop} on {Advances} in {Argumentation} in {Artificial} {Intelligence} 2022}, volume 3354 of \emph{{CEUR} {Workshop} {Proceedings}}. Udine, Italy: CEUR.
\newblock ISSN: 1613-0073.

\bibitem[{Dickie et~al.(2025)Dickie, Lauren, Belardinelli, Rago, and Toni}]{A_Dickie_2025}
Dickie, C.; Lauren, S.; Belardinelli, F.; Rago, A.; and Toni, F. 2025.
\newblock Aggregating bipolar opinions through bipolar assumption-based argumentation.
\newblock \emph{Autonomous Agents and Multi-Agent Systems}, 39(1).

\bibitem[{Dung(1995)}]{O_Minh_1995}
Dung, P.~M. 1995.
\newblock On the acceptability of arguments and its fundamental role in nonmonotonic reasoning, logic programming and n-person games.
\newblock \emph{Artificial Intelligence}, 77(2): 321--357.

\bibitem[{Lundberg and Lee(2017)}]{shap}
Lundberg, S.; and Lee, S.-I. 2017.
\newblock A Unified Approach to Interpreting Model Predictions.
\newblock arXiv:1705.07874.

\bibitem[{Marc~Feger(2025)}]{L_Feger_2025}
Marc~Feger, S.~D., Katarina~Boland. 2025.
\newblock Limited Generalizability in Argument Mining: State-Of-The-Art Models Learn Datasets, Not Arguments.
\newblock \emph{Proceedings of the 63rd Annual Meeting of the Association for Computational Linguistics (Volume 1: Long Papers)}, 23900--23915.

\bibitem[{Mensonides et~al.(2019)Mensonides, Harispe, Montmain, and Thireau}]{mensonides}
Mensonides, J.-C.; Harispe, S.; Montmain, J.; and Thireau, V. 2019.
\newblock Automatic Detection and Classification of Argument Components using Multi-task Deep Neural Network.
\newblock In \emph{{3rd International Conference on Natural Language and Speech Processing}}. Trento, Italy.

\bibitem[{Modgil and Prakken(2014)}]{aspic}
Modgil, S.; and Prakken, H. 2014.
\newblock The ASPIC+ framework for structured argumentation: a tutorial.
\newblock \emph{Argument \& Computation}, 5(1): 31--62.

\bibitem[{OpenAI(2025)}]{gpt41}
OpenAI. 2025.
\newblock Introducing GPT-4.1 in the API --- openai.com.
\newblock \url{https://openai.com/index/gpt-4-1/}.

\bibitem[{Peldszus and Stede(2015)}]{AMT1}
Peldszus, A.; and Stede, M. 2015.
\newblock An annotated corpus of argumentative microtexts.
\newblock In \emph{First European Conference on Argumentation: Argumentation and Reasoned Action}. Lisbon, Portugal.

\bibitem[{Peter~Potash(2017)}]{H_Potash_2017}
Peter~Potash, A.~R., Alexey~Romanov. 2017.
\newblock Here's My Point: Joint Pointer Architecture for Argument Mining.
\newblock \emph{Proceedings of the 2017 Conference on Empirical Methods in Natural Language Processing}, 1364--1373.

\bibitem[{Rahwan(2009)}]{Rahwan2009}
Rahwan, I. 2009.
\newblock \emph{Argumentation in Artificial Intelligence}.
\newblock Springer Science \& Business Media: Springer Science \& Business Media.
\newblock ISBN 9780387981970, 0387981977.

\bibitem[{Ribeiro, Singh, and Guestrin(2016)}]{ribeiro2016}
Ribeiro, M.~T.; Singh, S.; and Guestrin, C. 2016.
\newblock "Why Should I Trust You?": Explaining the Predictions of Any Classifier.
\newblock arXiv:1602.04938.

\bibitem[{Skeppstedt, Peldszus, and Stede(2018)}]{AMT2}
Skeppstedt, M.; Peldszus, A.; and Stede, M. 2018.
\newblock More or less controlled elicitation of argumentative text: Enlarging a microtext corpus via crowdsourcing.
\newblock In \emph{Proceedings of the 5th Workshop on Argument Mining, EMNLP 2018}. Brussels, Belgium.

\bibitem[{Slack et~al.(2021)Slack, Hilgard, Singh, and Lakkaraju}]{slack2021reliableposthocexplanations}
Slack, D.; Hilgard, S.; Singh, S.; and Lakkaraju, H. 2021.
\newblock Reliable Post hoc Explanations: Modeling Uncertainty in Explainability.
\newblock arXiv:2008.05030.

\bibitem[{Stab and Gurevych(2016)}]{AnnotatedEssays}
Stab, C.; and Gurevych, I. 2016.
\newblock Parsing Argumentation Structure in Persuasive Essays.
\newblock \emph{arXiv preprint arXiv:1604.07370}.

\bibitem[{Sutton(2019)}]{BitterLesson}
Sutton, R. 2019.
\newblock The Bitter Lesson.
\newblock \url{http://www.incompleteideas.net/IncIdeas/BitterLesson.html}.

\bibitem[{Thia@Microsoft(2025)}]{ZeroTrustMS2025}
Thia@Microsoft. 2025.
\newblock Zero-Trust Agents: Adding Identity and Access to Multi-Agent Workflows | Microsoft Community Hub.
\newblock \url{https://techcommunity.microsoft.com/blog/azure-ai-foundry-blog/zero-trust-agents-adding-identity-and-access-to-multi-agent-workflows/4427790}.

\end{thebibliography}

\end{document}